\pdfoutput=1

\documentclass[11pt]{article}

\usepackage{ACL2023}

\usepackage{times}
\usepackage{latexsym}
\usepackage{amsmath}
\usepackage{graphicx}
\usepackage{bbm}
\usepackage[T1]{fontenc}

\usepackage[utf8]{inputenc}

\usepackage{microtype}

\usepackage{caption}
\usepackage{subcaption}
\usepackage{pgfplots}
\usetikzlibrary{patterns}
\usepackage{filecontents}
\usepackage{algorithmic,algorithm}
\usepackage{csvsimple}
\usepackage{amsmath}

\usepackage{inconsolata}

\usepackage{booktabs} 
\usepackage{multirow}
\usepackage{xcolor}
\usepackage{xcolor, soul}
\sethlcolor{pink}
\usepackage{amssymb}

\usepackage[textsize=scriptsize]{todonotes}

\title{A Critical Evaluation of Evaluations for Long-form Question Answering
}

\author{Fangyuan Xu$^{\diamondsuit}$\thanks{$^{*}$Equal contribution.}~~  Yixiao Song$^{\heartsuit*}$ ~~ Mohit Iyyer$^{\heartsuit}$ ~~ Eunsol Choi$^{\diamondsuit}$ \\
$^\diamondsuit$The University of Texas at Austin, $^\heartsuit$University of Massachusetts Amherst \\
 \hspace{0.5em} {\texttt{\{fangyuan, eunsol\}@utexas.edu}} \\
 {\texttt{yixiaosong@umass.edu, miyyer@cs.umass.edu}}
 }

\begin{document}
\maketitle
\begin{abstract} 
Long-form question answering (LFQA) enables answering a wide range of questions, but its flexibility poses enormous challenges for evaluation. We perform the first targeted study of the evaluation of long-form answers, covering both human and automatic evaluation practices. We hire domain experts in seven areas to provide preference judgments over pairs of answers, along with free-form justifications for their choices. We present a careful analysis of experts' evaluation, which focuses on new aspects such as the comprehensiveness of the answer. Next, we examine automatic text generation metrics, finding that no existing metrics are predictive of human preference judgments. However, some metrics correlate with fine-grained aspects of answers (e.g., coherence). We encourage future work to move away from a single ``overall score'' of the answer and adopt a multi-faceted evaluation, targeting aspects such as factuality and completeness. We publicly release all of our annotations and code to spur future work into LFQA evaluation.\footnote{{\url{https://github.com/carriex/lfqa_eval}}}

\end{abstract}

\section{Introduction}
{Long-form question answering}~\cite[][henceforth LFQA]{Fan2019ELI5LF, Krishna2021HurdlesTP, nakano2021webgpt, su-etal-2022-read}, an emerging research area within QA, requires  systems to \emph{generate} long and complex answers to questions by leveraging large language models and evidence document retrievers. While remarkable strides have been made in LFQA model development, the current state of LFQA \emph{evaluation} is dire: most prior papers use a combination of crowdsourced human annotations and simple string-matching metrics (e.g., ROUGE). We present the first study of the evaluation of long-form answers, exploring both human and automatic evaluation protocols to better understand how we should evaluate LFQA moving forward.




\noindent \textbf{Human evaluation:} In most prior human LFQA evaluations~\citep{Krishna2021HurdlesTP,nakano2021webgpt}, crowd annotators are given a question, two candidate answers, and (optionally) evidence documents, and they are asked to identify the better answer. However, crowdworkers do not necessarily have the expertise or background knowledge to reliably judge properties such as factuality~\citep{gillick-liu-2010-non,iskender-etal-2020-best}. Thus, we hire \emph{domain experts} in seven different fields (e.g., biology, economics) to perform the same answer preference task and additionally provide detailed justifications as to why they chose a particular answer. Analyzing their justifications reveals that experts consider properties such as completeness and factuality to be more decisive than surface-level aspects (e.g., conciseness and level of detail) on which crowdworkers tend to fixate. Additionally, even experts often disagree with each other about which answer is better; this disagreement stems from valuing fine-grained answer properties differently.

\begin{table*}[]
\scriptsize
\resizebox{0.98\textwidth}{!}{%
\begin{tabular}{@{}c@{\hspace{1.3ex}}p{2cm}@{\hspace{2ex}}p{3.7cm}@{\hspace{2ex}}p{3.7cm}@{\hspace{2ex}}p{3.2cm}@{}}
\toprule
\textbf{Aspect} &
  \textbf{Question} &
  \textbf{Answer A} &
  \textbf{Answer B} &
  \textbf{Expert Justification} \\\midrule
\multirow{1}{*}{\rotatebox[origin=c]{90}{\bf Factuality }} &
  Does your body absorb all blood clots? How does it do it? Does location affect it? &
  \emph{(human)} Kind of. Your blood will dissolve clots over time. I've got a clotting disorder and \textbf{need to take thinners to make that happen faster}, but for most they break down clots naturally [...]   &

  \emph{(model)} No, your body does not absorb all blood clots. Clots are usually broken down by the body's natural clotting mechanism, but some clots may persist and cause problems. Location can affect how quickly a clot is broken down [...] &
  \emph{(expert)} {[}...{]} Answer A mentions some \hl{inaccurate} and irrelevant information, \hl{like role of blood thinner which is not correct as they cannot break down the clots}. {[}...{]}\\\midrule
  \multirow{1}{*}{\rotatebox[origin=c]{90}{\bf Completeness}} & Why do people go bald on the top of their head and still have some on the sides and not vice-versa? & \emph{(model)} 
  \textbf{One reason} is that the hair follicles on the top of the head are more sensitive to a hormone called dihydrotestosterone (DHT). DHT is [...] found in both men and women. [...] \textbf{Another reason} is that the hair on the sides and back of the head is not as exposed to the sun and other environmental factors, [...] & \emph{(human)} 
  \textbf{Seems unsettled but here’s a theory}: The most recent hypothesis suggests that the hair loss process begins during puberty, when growth of the skull and the muscles in the forehead and neck increases the tension in a tight band of tissue stretching over the top of the head. The more DHT (a type of testosterone) there is... &
  \emph{(expert)} \hl{Answer A is the better choice as it describes both the hormonal and environmental causes and Answer B only focuses on one theory which might not be 100 percent accurate.} [...] According to research, baldness is due to genes. In 95 percent cases, balding is due to androgenetic alopecia [...] 
   \\
  \bottomrule
\end{tabular}%
}
\vspace{-0.6em}
\caption{Examples of two fine-grained aspects, \textbf{factuality} (top) and \textbf{completeness} (bottom), that were decisive factors in our expert annotators' preference of one answer over another. The \textit{human} answers are from the \href{https://www.reddit.com/r/explainlikeimfive/}{r/explainlikeimfive} subreddit and the \textit{model} answers are generated zero-shot by \texttt{text-davinci-002}. See Table \ref{tab:takeaway} for more examples.}
\label{tab:maintakeaway}
\vspace{-0.15in}
\end{table*}

\noindent \textbf{Automatic evaluation:} As human evaluation is slow and expensive, developing a reliable automatic LFQA evaluation metric is crucial for speeding up model development. While ROUGE \cite{lin-2004-rouge} has been shown to be misleading for LFQA~\cite{Krishna2021HurdlesTP,Wang2022ModelingEI}, do any other existing text generation metrics correlate to human judgments of answer quality? Can we train a metric to mimic human preference judgments? To answer these questions, we curate a suite of 12 automatic metrics and measure how they correlate to human judgments of both ``overall quality''  and two fine-grained aspects (coherence and faithfulness). None of these metrics reliably matches human judgments of overall answer quality. However, automatic metrics such as QAFactEval \cite{fabbri-etal-2022-qafacteval} and RankGen \cite{krishna2022rankgen} show potential at modeling fine-grained aspects of LFQA answers, which can spur research on a new generation of automatic LFQA metrics. 

Overall, we provide the first thorough study of LFQA evaluation and shed light on the components of good long-form answers. As part of our exploration, we collected and will release a small-scale dataset of expert evaluation of long-form answers (260 ratings and justifications over 140 answer pairs). We conclude by providing recommendations for the future of human and automatic LFQA evaluation, encouraging the community to hire expert evaluators and move from poorly-defined judgments of ``overall preference''  to a multi-faceted evaluation modeling attributes such as answer completeness, factuality, and ease of understanding.





\section{Background and related work}\label{sec:background}

We begin by reviewing the evaluation protocols used by prior work in LFQA, which has centered around a dataset scraped from the “Explain Like I’m
Five” subreddit~\citep[][ELI5]{Fan2019ELI5LF}.\footnote{\url{https://www.reddit.com/r/explainlikeimfive}} We include brief review of evaluation in other text generation tasks in Appendix~\ref{subsec:related_work}.

\paragraph{Prior automatic evaluations:}
 Early work on LFQA \cite{Fan2019ELI5LF} uses ROUGE~\citep{lin-2004-rouge} to measure the similarity of human reference answers to model-generated answers.~\citet{Krishna2021HurdlesTP} find that ROUGE is not a meaningful metric due to the open-ended nature of long-form answers, but they do not examine other automatic metrics. Given the difficulty of evaluation, recent works re-scoped the task to allow more reliable evaluation: \citet{Wang2022ModelingEI} focus on exemplification in long-form answers by treating this sub-task as a retrieval problem, while \citet{stelmakh2022asqa} aim to evaluate long form answers limited to ambiguous factoid questions that cover the different disambiguated questions and their corresponding answers. However, these evaluation protocols cannot be easily adapted to the general LFQA task: the metric in \citet{stelmakh2022asqa}, for example, requires a list of disambiguated questions and their answers, which is not available for many questions. 
 

\vspace{-0.1in}

\paragraph{Prior human evaluations:}\label{sec:prev_human_eval}
We summarize the human evaluation studies conducted by two previous studies, \textsc{Hurdles} \cite{Krishna2021HurdlesTP} and \textsc{WebGPT} \cite{nakano2021webgpt}. Both works evaluate via A/B testing (i.e., choose which of two candidate answers is better), and they collected judgments of overall answer quality, factuality, and coherence. While both works recruited non-expert annotators and collect only one-way annotations, \textsc{WebGPT}'s evaluation allows annotators to look at a set of evidence documents when judging the answer, and they also collect optional free-form justifications from the annotators to justify their choice. While fine-grained aspects such as coherence \cite{goyal2022snac,jiang+al.naacl22} and factuality \cite{goyal-durrett-2020-evaluating, Laban2022SummaCRN} have been studied before for other tasks such as summarization, ours is among the first works to study LFQA-centric properties such as completeness or ease of understanding.

\section{How do domain experts evaluate long-form answers?}\label{sec:annotations}


\begin{table}[]
\scriptsize
\resizebox{\columnwidth}{!}{%
\begin{tabular}{@{}l@{\hspace{-0.2ex}}rrr@{\hspace{-0.2ex}}r@{}}
\toprule
\textbf{Category} &   \multicolumn{2}{c}{\textbf{Preference}} & \textbf{Fleiss'} \\
{\small (\# of experts)} &   {\small Upvote $\uparrow$ (H/H)} & {\small Model (H/M)} &   \textbf{$\kappa$}   \\ \midrule
Biology (3)     &   76.7\% & 53.3\%  & 0.52 \\
Physics (2)     &   50\% & 65\%  & 0.50 \\
Chemistry (1)   &   70\% & 50\%  & --   \\
Economics (2)   &   60\% & 90\%  & 0.40 \\
Law (1)         &   60\% & 90\%  & --   \\
Tech/CS (1)     &   40\% & 60\%  & --   \\
History (3)     &   80\% & 24.4\%  & 0.65 \\\midrule
\textbf{Average}         &   62.4\% & 61.8\% & -- \\\bottomrule
\end{tabular}%
}\vspace{-0.4em}
\caption{Results of our expert annotation of seven domains, where the two candidate answers are either both human-written (H/H) or human-written vs.\ model-generated (H/M).
We report how often the highly-upvoted answer was preferred in H/H, and how often the model-generated answers are preferred in H/M.}
\label{tab:expert-number}
\vspace{-0.15in}
\end{table}

Prior LFQA human evaluations use non-expert crowdworkers to evaluate highly domain-specific answers, either with no access to external information~\citep{Krishna2021HurdlesTP} or access to only model-retrieved evidence documents~\citep{nakano2021webgpt}. Both settings are problematic: non-experts cannot be relied on to judge the correctness of answers in isolation, and they also cannot be expected to thoroughly comprehend evidence documents and judge their validity or relevance to the answer~\citep{Gao2022AttributedTG}. While~\citet{nakano2021webgpt} solicit optional free-form justifications from their workers to explain their preference judgments, it remains unclear how well these workers can judge \emph{correctness} in fields that are not their expertise. Our first contribution is to hire \emph{domain experts} in seven fields (see Table \ref{tab:expert-number}) and have them evaluate both human-written and model-generated answers via A/B judgments as well as paragraph-length free-form justifications. An analysis of the expert annotations reveals a complex and subjective interplay between many different fine-grained aspects of LFQA answers (e.g., completeness, factuality) that pose challenges for future LFQA evaluation.


\subsection{Collecting expert judgments} 



 \paragraph{Hiring experts:} We recruit domain experts on the freelancing platform Upwork for seven domains shown in Table \ref{tab:expert-number}. Each expert has earned at least a bachelor's degree in the target domain and has expertise performing tasks in that domain (e.g., summarizing scientific articles or being a teacher of the domain). As shown in Table \ref{tab:expert-number}, we hire 1-3 experts per domain. Given a question and two candidate answers, the experts were asked to choose which of the answers is better (\emph{overall preference}), indicate whether the decision was difficult to make (e.g., because both answers were of similar quality), and lastly to justify their choice in a free-form paragraph. The evaluation tasks are hosted on Label Studio.\footnote{Figure~\ref{fig:annotation_interface} contains a screenshot of our annotation interface. \url{https://labelstud.io/} } The experts reported that they spent 15 to 30 minutes per question, which shows the demanding nature of the annotation task. We accordingly paid \$$3.25$ per question, which resulted in a total cost of \$845 to collect 260 expert judgements.\footnote{We explain in Appendix \ref{sec:expert_imp_details} why the numbers of experts in each domain differ.}

\vspace{-0.05in}

\paragraph{Setting up the A/B task:}

Following prior work, we conduct A/B preference testing on two answers to the same question. We include two settings: (1) H/M: comparing a model-generated answer with a highly-upvoted human-written answer, and (2) H/H: comparing a highly-upvoted human-written answer to an answer with fewer upvotes (where upvotes are a noisy proxy to answer quality).\footnote{Reddit users give upvotes to content to show their support or approval of the content.} The first setting is intended to identify common classes of errors made by state-of-the-art LFQA systems, while the second setting is more of a sanity check exploring whether low-effort human answers make similar errors to models.

We chose GPT-3 \texttt{text-davinci-002} model (175B)~\cite{Brown2020LanguageMA} as the LFQA model to evaluate. A small-scale qualitative analysis found that zero-shot GPT-3 possesses more advanced LFQA capabilities than fine-tuned LFQA systems built on smaller language models. Since this model may have already seen the entire ELI5 dataset released by~\citet{Fan2019ELI5LF} during its pretraining, we scrape more recent questions from the \href{https://www.reddit.com/r/explainlikeimfive/}{r/explainlikeimfive} and \href{https://www.reddit.com/r/AskHistorians/}{r/AskHistorians} subreddits posted between July to December 2021.\footnote{\texttt{text-davinci-002} was trained on data up to June 2021.} Question askers on the ELI5 subreddit often categorize their questions into domains via the \texttt{flair} label, which enables us to perform a domain-specific analysis.\footnote{The details on domain identification are in Appendix \ref{sec:expert_imp_details}.} We randomly sample 20 questions per domain except for the history domain, which has 15 questions in the H/M setting and 5 in H/H. This discrepancy is due to the difficulty of finding history questions with a moderate answer length. As shown in Figure \ref{fig:human-model-ans-len} and Table \ref{tab:length_stats}, human-written answers to history questions are much longer than the answers in the other domains, even after careful screening.

To obtain model-generated answers, we prompt the model in a zero-shot manner with the following prompt: ``Generate a long answer to the following question with examples and references when necessary.'' For decoding, we used the default decoding setup in the API (i.e., top $p=1$ and temperature$=0.7$). 


\vspace{-0.05in}

\subsection{Quantitative results}


As shown in Table~\ref{tab:expert-number}, experts surprisingly display a slight preference (61.8\%) for \emph{model-generated} answers from GPT-3 compared to human answers; as a sanity check, they exhibit preference (62.4\%) for highly-upvoted human answers over those with fewer upvotes. The preference of our annotators for model-generated answers is corroborated by similar findings for summarization by~\citet{Liu2022RevisitingTG}, who show that GPT-3 generated summaries score higher than reference summaries. 

\begin{figure}[t]
    \centering
    \includegraphics[width=0.48\textwidth]{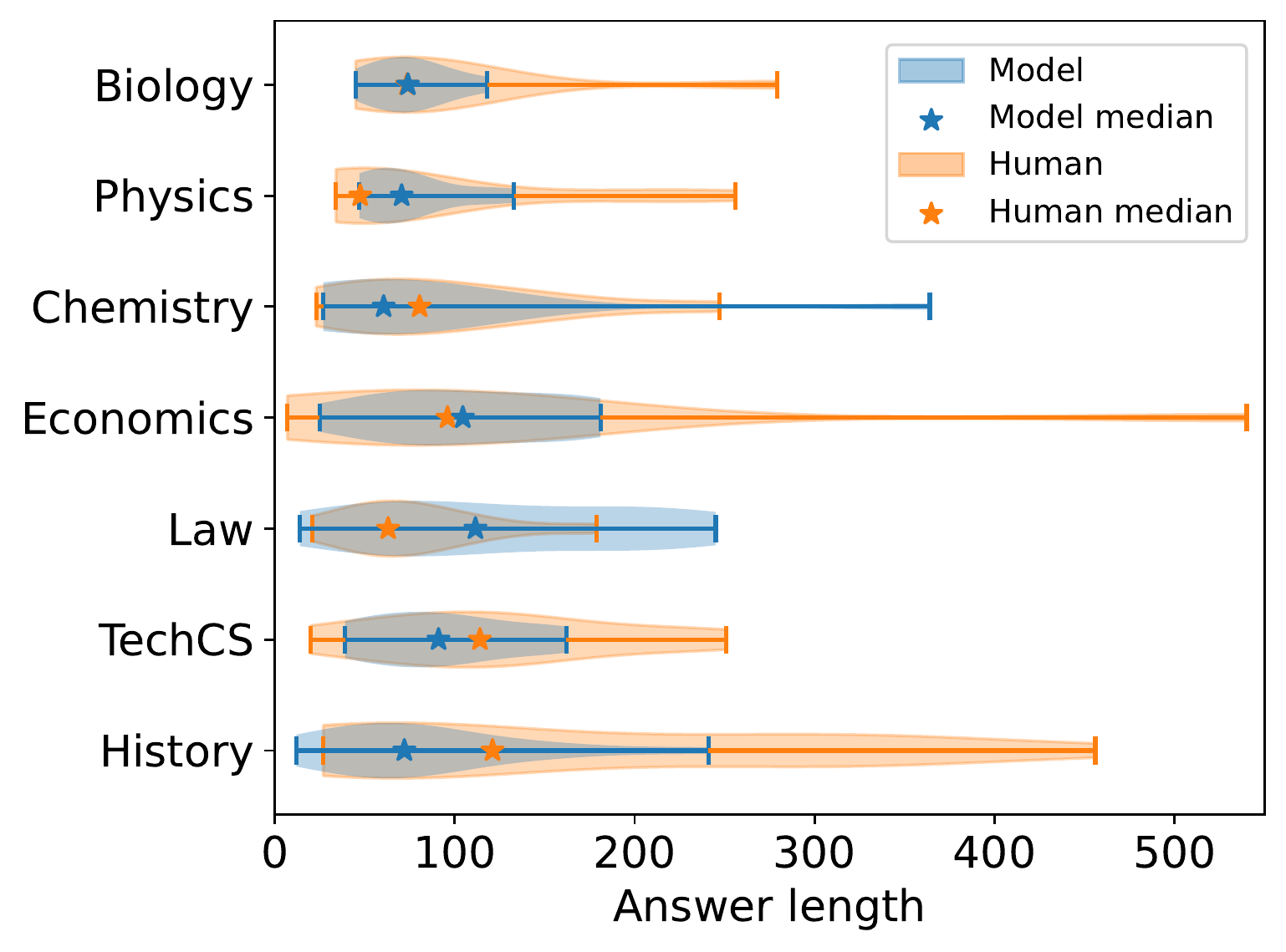}
    \vspace{-0.25in}
    \caption{Answer length distribution in the comparison of model-generated and human-written answers (H/M) in our expert-annotated dataset. History is the hardest domain for models and also has the largest discrepancy between model and human answer length. There are 75 questions and 75 human-written and model-generated answers.
    }
    \label{fig:human-model-ans-len}
    
\end{figure}

Comparing different domains, we observe that model-generated answers are strongly preferred in economics (90\%) and law (also 90\%), while human answers are preferred in the history domain (75.6\%). To understand the divergence in preferences for different domains, we report the answer length distribution of both answer types in the H/M setting in our expert-annotated dataset in Figure \ref{fig:human-model-ans-len}. The model's struggles in history domain are likely because this domain contains the longest and most complex {questions} as well as human answers (averaging 356 words long {in the H/M setting}) out of all domains. Table \ref{tab:length_stats} in the appendix report the length of questions, model-generated, and human-written answers of the whole expert-annotated dataset.

\vspace{-0.05in}

\paragraph{Expert (dis)agreement:} We report Fleiss' $\kappa$~\cite{fleiss1971measuring,landis1977measurement,fleiss2013statistical} as a measure of agreement in Table \ref{tab:expert-number}. Our expert A/B testers achieved fair agreement in economics, moderate agreement in biology and physics, and a substantial agreement in history. We observe that agreement increases when comparing a high and low-upvoted human answer together, as opposed to comparing model-generated answers with human answers. We emphasize that disagreement is \emph{not} a failure of one of the experts to properly evaluate the answers. In fact, disagreement within experts highlights the challenges (and futility) of judging ``overall answer quality'' in this way. There are many salient properties of long-form answers, which we discuss next, and deciding how to value each property when coming up with an overall preference is highly subjective (see Appendix Table~\ref{tab:comment_examples} for several examples). 

\subsection{What makes one answer better than another?}\label{subsec:manual-analysis}

\begin{figure*}[t]
    \centering
    \includegraphics[width=\textwidth]{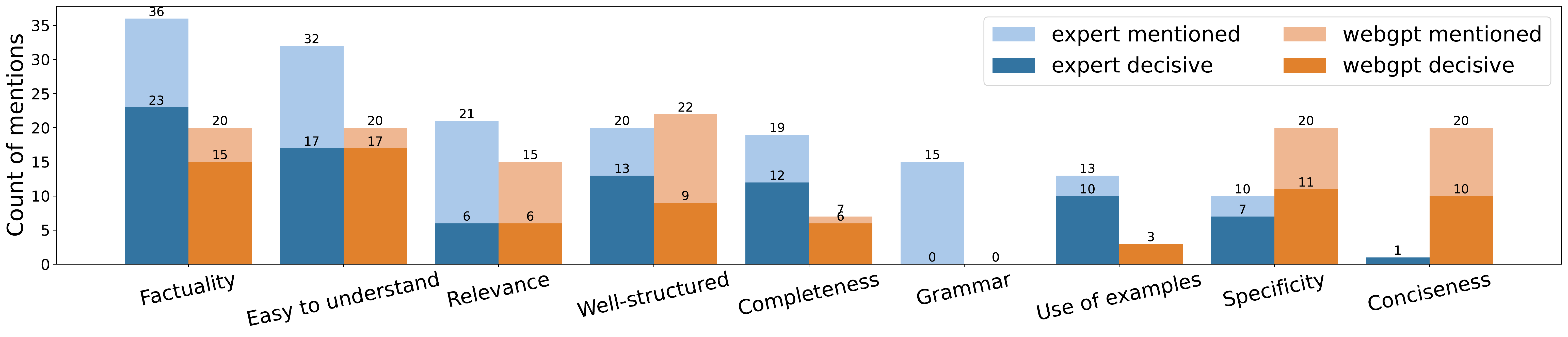}\vspace{-0.5em}
    \caption{We manually analyzed 50 justifications each from both experts and \textsc{WebGPT} crowd annotators. We report nine frequently-mentioned fine-grained aspects here. The plot shows that experts and crowdworkers disagree on which aspects are more decisive, and that experts are more sensitive to factuality and completeness.}
    \label{fig:comment_analysis}
    \vspace{-0.05in}
\end{figure*}

To better understand the various components of a good long-form answer, we perform an analysis on the free-form justifications  collected from both our expert annotators as well as \textsc{WebGPT} crowd annotators from \citet{nakano2021webgpt}. \textsc{WebGPT} allowed \emph{optional} justifications, and many of them are not very long or detailed. Our justification is about three times longer on average (statistics can be found in Table \ref{tab:criqitue_stats} in the Appendix). Our analysis focuses on the model-generated vs.\ human-written answer setting, where the model is either zero-shot GPT-3 (our work) or the 175B \textsc{WebGPT} model.  
Concretely, we analyze 50 randomly sampled justifications from each population. Our analysis is limited in that these two comparisons do not consider the same set of questions. We identify and code nine fine-grained aspects that are mentioned in them, and mark whether these aspects are decisive factors for making the preference judgment. The results are summarized in Figure \ref{fig:comment_analysis}, and we highlight takeaways below. 

%



\vspace{-0.05in}

\paragraph{Experts are better judges of factuality:} Perhaps unsurprisingly, our experts mention \textbf{factuality} in their justifications almost twice as frequently as crowdworkers (36 to 20), and it is the most common aspect referenced by experts. As an example, in the first row of Table \ref{tab:maintakeaway}, the expert accurately points out incorrect information in Answer A about blood thinners breaking up clots. Since \textsc{WebGPT} annotators lack domain expertise, they  generally judge factuality by checking if a statement is supported in evidence documents, which gives them only limited coverage over the full answer. 

\vspace{-0.05in}

\paragraph{Experts value answer completeness:} We observe that experts mention \textbf{completeness} as a decisive criteria twice as often than \textsc{WebGPT} annotators (12 vs.\ 6). Completeness refers to whether the answer adequately addresses all aspects of the question or provides all necessary information to clarify the question. Judging completeness requires deeper domain expertise than a handful of retrieved articles offer. {As an example,} in the second row of Table~\ref{tab:maintakeaway}, the expert states that Answer B mentions only one reason why people go bald (hormonal), while Answer A mentions hormonal and environmental factors and is thus superior.\footnote{The expert further points out that both answers miss a third major cause of baldness: genetics.}

\vspace{-0.05in}

\paragraph{All annotators value ease of understanding.} Both experts and crowdworkers mention \textbf{easiness to follow} as a decisive criterion at the same frequency; in fact, this is the most decisive aspect for both populations. One of the main goals of LFQA is to convey the answer of a question to a non-expert; as such, it makes sense that this property is so critical. We emphasize that this has \emph{never} been evaluated in prior LFQA research and encourage future work to embrace it as a major component.

\vspace{-0.05in}

\paragraph{Non-experts focus on surface-level properties:} \textsc{WebGPT} annotators are far more likely to mark \textbf{conciseness} and \textbf{specificity} as decisive factors for their preferences than experts. They prefer shorter to-the-point answers, despite the fact that such answers might be incomplete, and they also prefer answers that include specific details instead of generalities. We note that these properties are much more feasible to judge for crowdworkers than factuality and completeness, which is likely a reason why they are mentioned so frequently (Table \ref{tab:takeaway} in the appendix for examples).

\subsubsection{Do models understand justifications of human preferences?}
Our manual analysis of the justifications shows that experts consider a wide range of aspects when forming their decision. 
Detailed justifications of generated answers are useful in understanding why an answer was preferred, but they are costly to obtain. Generating these justifications automatically and evaluating them is outside the scope of this paper. Instead, we perform a simpler evaluation via a proxy task: given a justification with masked references to both candidate answers, can a model disambiguate the missing references? An example of the task is below:

\begin{quote}
\small
    \textbf{Input}: Question: $q$ Answer A: $a_{1}$ Answer B: $a_{2}$ Comment: Both answers are coherent, but Answer <$\tt{extra\_id\_0}$> is completely irrelevant to the question since it is about a bionic ear instead of a person learning speech when they get a hearing implant. Answer <$\tt{extra\_id\_1}$> is relevant and a complete, concise answer. \\ 
    \textbf{Expected Output}: <$\tt{extra\_id\_0}$> B <$\tt{extra\_id\_1}$> A 
\end{quote}



\begin{table}
\small
\begin{center}
\begin{tabular}{@{}llccc@{}}
\toprule
\textbf{Data} &\textbf{Model} & \multicolumn{3}{c}{\textbf{Token level EM}}\\
& & O$\uparrow$ & F$\downarrow$ & R \\
\midrule
Expert  & T5-base & 0.36 & 0.37 & 0.33 \\ 
& T5-large & 0.51 & 0.44 & 0.41 \\ 
& T5-3B & 0.66 & 0.36 & 0.48\\ 
& T5-11B & \textbf{0.76} & \textbf{0.28} & 0.47\\
\midrule
\textsc{WebGPT} & T5-base & 0.40 & 0.38 & 0.37 \\ 
& T5-large & 0.50 & 0.49 & 0.50 \\ 
& T5-3B & 0.60 & 0.46 & 0.53 \\
& T5-11B & \textbf{0.65} & \textbf{0.40} & 0.54 \\
\bottomrule
\end{tabular} 
\end{center}\vspace{-0.3em}
\caption{Results on masked justification reference prediction: \textbf{O}riginal comments, \textbf{F}lipped comments and \textbf{R}andom comments. The larger LMs can identify references in justifications better.}
\label{tab:full_masked_critique_results}
\end{table}




We experiment with pretrained T5 checkpoints ~\cite{t5} of different sizes (220M, 770M, 3B, and 11B parameters) on our task zero-shot.\footnote{We experimented with two-shot prompting with GPT-3 but observed worse results compared to the outputs from T5-3B and T5-11B, potentially because the task resembles the pretraining setup of T5.} For each (question $q$, answer pairs ($a_{1}$, $a_{2}$), justification $j$), we construct three types of inputs: \textbf{Original}: The original justification $j$ with $(q, a_{1}, a_{2})$, \textbf{Flipped}: The original justification $j$ with flipped answer identity $(q, a_{2}, a_{1})$,  \textbf{Random:} $j$ with randomly paired $q', a_{1}', a_{2}'$, as a baseline. We evaluate using token-level exact match, which gives the model credit only when its output exactly matches that of the target. We expect better than random performance on \textbf{Original} and worse than random performance on \textbf{Flipped} if the model comprehends the justifications. 

Results are shown in Table \ref{tab:full_masked_critique_results}. We see that T5-3B an T5-11B are able to comprehend the justifications, as they show different results for original and perturbed comments. This suggests adapting LMs for multi-faceted automatic evaluations of long-form answers is promising. Preprocessing details on this study are described in Appendix~\ref{subsec:full_masked_critique_anlaysis}





\vspace{-0.05in}

\section{Do automatic metrics correlate with human judgments?} \label{sec:metrics}

The experiments in the previous section establish that LFQA is very difficult for humans to converge on in terms of an ``overall'' score, as even domain experts disagree with each other when choosing a ``better'' LFQA answer. Furthermore, several properties of these answers are important to evaluate, including factuality, relevance, and coherence, among others. Do existing automatic text generation metrics correlate with human judgments of these fine-grained aspects, or``overall'' answer preference? We now explore this question with a wide range of text generation evaluation metrics. 

\subsection{Text generation metrics}\label{sec:automatic_metrics}
We experiment with existing text generation metrics and metrics that we train directly on the human preference judgments.


\subsubsection{General-purpose generation metrics}
Prior work used existing text generation metrics (e.g., ROUGE) to evaluate LFQA. The metrics were initially designed for other text generation tasks  (e.g., translation or summarization), and their usage has not been validated for LFQA. 

\vspace{-0.05in}

\paragraph{Reference-based metrics:} Many generation metrics assume access to human-written references (in our case, gold answers), which are used to compute similarity scores to model-generated text. 
Of these, we evaluate \textbf{ROUGE}~\cite{lin-2004-rouge}, which is the only reference-based evaluation metrics employed by prior work for LFQA, as well as \textbf{BERTScore}~\cite{zhang2019bertscore} and \textbf{BLEURT}~\cite{sellam2020bleurt}, which leverage pretrained language models and have shown to be effective in evaluating many generation tasks~\citep{kasai-etal-2022-bidimensional}. A major limitation of reference-based metrics for LFQA is the huge space of valid output answers for any given question, which has been noted in prior work~\citep{Wang2022ModelingEI}.




\vspace{-0.05in}

\paragraph{Answer-only metrics:}
Some aspects, such as fluency and coherence, can be determined by looking at just the answers alone. Thus, we also examine a set of answer-only automatic metrics: (1) \textbf{Self-BLEU}~\cite{zhu2018texygen}, which measures the diversity of generated text (higher scores mean lower diversity) and has been previously used in open-ended generation \cite{holtzman2019curious}; and (2) \textbf{GPT-2 perplexity}, which prior work on constrained generation~\cite{zhang-etal-2020-language-generation,qin2022cold} has used to evaluate fluency.


\vspace{-0.05in}

\paragraph{(Question, answer) metrics:} Good answers should be \textit{relevant} to the question asked, so we can model $p(q|a)$ to rank answers using the following methods: (1) \textbf{Zero-shot question likelihood}, which uses the instruction-tuned T0 model~\cite{sanh2022multitask} to calculate the likelihood of the question given the long-form answer; (2) \textbf{BARTScore}~\citep{bart-score}, which is an encoder-decoder model fine-tuned on text summarization; and (3) \textbf{RankGen}~\cite{krishna2022rankgen}, which is an encoder model trained contrastively to score model-generated sequences (in our case, answers) given a prefix (the question).


\vspace{-0.05in}

\paragraph{(Answer, evidence) metrics:} Arguably the most challenging aspect of LFQA evaluation is to measure the correctness of the answer. While there are no existing factuality metrics for LFQA, the task is related to faithfulness in summarization. Metrics for faithfulness assume access to a set of evidence documents and evaluate whether a text is supported by the evidence~\cite{kryscinski-etal-2020-evaluating, goyal-durrett-2020-evaluating, barrantes2020adversarial, Laban2022SummaCRN}. We experiment with the \textbf{QAFactEval} metric~\cite{fabbri-etal-2022-qafacteval}, which evaluates faithfulness by comparing answers from the summary (in our case, the answer) and the evidence document (retrievals from the \textsc{WebGPT} LFQA system).




\subsubsection{Trained LFQA metrics} \label{sec:reward_model}
The metrics discussed so far are not trained on long-form answers. We now shift to training an LFQA evaluation metric directly on human-annotated preference judgments of pairs of long-form answers. Prior work from OpenAI~\cite{nakano2021webgpt} experimented with learning an evaluation metric by fine-tuning \textsc{WebGPT} to rank pairs of answers. As this model is not publicly available, we fine-tune a smaller-scale pretrained language model (176M Longformer-Base model) and rely on OpenAI's API to fine-tune bigger pretrained language model (6B GPT3 \texttt{text-curie-001} model.\footnote{To the best of our knowledge, OpenAI has not clarified the exact size of each of the models in the API. We use this estimation:\url{https://blog.eleuther.ai/gpt3-model-sizes/}.}) {Details of fine-tuning setup are in Appendix~\ref{subsec:finetuningdetails}.} 




\vspace{-0.05in}

\paragraph{Data}


We use comparison data collected by \citet{nakano2021webgpt} for fine-tuning, which contains 17,598 preference annotations. We remove ties and randomly split the data into train, validation and test sets with a 70\%, 15\%, 15\% ratio. More details are provided in Appendix Table \ref{tab:rm_data_stats}.


\vspace{-0.05in}

\paragraph{Fine-tuning Longformer} Our learned metric \textit{f} takes in question $q$, answer $a$, and optionally evidence documents $d$ to produce a scalar score. We encode [\textit{q}, \textit{a}] and [\textit{a}, \textit{d}] separately with an encoder model and concatenate respective \texttt{[CLS]} representation then pass it to a linear layer to obtain a scalar score $s$. As our input text is relatively long, we fine-tune a Longformer encoder~\cite{Beltagy2020LongformerTL}. 

Following~\citet{nakano2021webgpt}, we train the model with cross-entropy loss such that the scores produced by $f$ rank a pair of answers ($a_{1}$,$a_{2}$) in the same order as the human preference. We estimate the likelihood that $a_{1}$ is preferred over $a_{2}$ as $ \frac{exp(s_{1})}{exp(s_{1}) + exp(s_{2})}$ where $s_{1} = f(q, a_{1}), s_{2} = f(q, a_{2})$. Given a set of answer pairs with gold preference $\hat{p}$, the loss is,

\vspace{-0.15in}

\begin{footnotesize}
\[ L = -(\mathbbm{1}[\hat{p}=a_{1}]logP(p=a_{1}) \\ + \mathbbm{1}[\hat{p}=a_{2}]logP(p=a_{2})), \]
\end{footnotesize}

\vspace{-0.05in}

\noindent where $\mathbbm{1}$ is the indicator function. We consider two inference settings, \textbf{longformer(D)}, which considers evidence documents, and \textbf{longformer} which takes the concatenation of [\textit{q}, \textit{a}] and [\textit{a}], as evidence documents are not always available.



\vspace{-0.05in}

\paragraph{Fine-tuning GPT-3}
To leverage the advanced capabilities of larger-scale language models, we use OpenAI API to finetune GPT-3 \texttt{text-curie-001} with the same comparison data split we used for the Longformer. Given a prompt consisting of question $q$, answer $a_{1}$ and answer $a_{2}$, the model is fine-tuned to output the label \texttt{Answer1} or \texttt{Answer2}. This metric takes a \textit{pair} of answers as input and outputs a preference, unlike the Longformer model which produces a score given a single answer.

\begin{table*}[ht]
\footnotesize
\begin{center}
\begin{tabular}{@{}lccccccccccc@{}}
\toprule
& \multicolumn{5}{c}{{\textbf{Overall}}} & \multicolumn{3}{c}{{\textbf{Coherence}}} & \multicolumn{3}{c}{{\textbf{Factuality}}}\vspace{0.1cm}\\
\midrule
\textbf{Data source} & Expert & \multicolumn{2}{c}{\textsc{WebGPT}} & \multicolumn{2}{c}{\textsc{Hurdles}} &   {\textsc{WebGPT}}  &   \multicolumn{2}{c}{\textsc{Hurdles}} &   {\textsc{WebGPT}}  &   \multicolumn{2}{c}{\textsc{Hurdles}} \\
   Setting & & \textbf{h/m} & \textbf{m/m} & \textbf{h/m} & \textbf{m/m} & \textbf{h/m} & \textbf{h/m} & \textbf{m/m} & \textbf{h/m} & \textbf{h/m} & \textbf{m/m} \\

\# pairs & 129 & 637 & 1,923 & 419 & 370 & 496  & 164 & 194 & 149 & 151 & 169 \\
\midrule
\multicolumn{11}{l}{\textit{\textbf{Baselines}}} \\
Random & 0.50 & 0.50 & 0.49 & 0.50 & 0.48 & 0.50 & 0.51 & 0.50 &  0.50 & 0.50 & 0.49  \\
Always Human & - & 0.61 & - & 0.81 & - & 0.70 & 0.87 & - &  0.52 & 0.95 & -\\
Length & 0.68& 0.52 & 0.57 & 0.61 & 0.48 & 0.38 & 0.62 & 0.49 & 0.57 & 0.68 & 0.57 \\ 
\midrule
\multicolumn{11}{l}{\textit{\textbf{Reference-based metrics}}} \\
ROUGE & 0.58$^{\dagger}$ & 0.53 & 0.53 & 0.43 & 0.52 & 0.54 & 0.46 & 0.48 & 0.46 & 0.40 & 0.51  \\ 
BERTScore & 0.57$^{\dagger}$ & 0.57 & 0.51 & 0.46 & 0.61 &  \textbf{0.62} & 0.39 & \textbf{0.69} & 0.48 & 0.39 & 0.61 \\ 
BLEURT & 0.62$^{\dagger}$ & 0.52 & 0.54 & 0.42 & 0.56 & 0.55 & 0.32 & 0.45 & 0.52 & 0.33 & 0.53 \\ 
\midrule
\multicolumn{11}{l}{\textit{\textbf{Answer-only metrics}}} \\
Self-bleu & 0.36 & 0.50 & 0.45 & 0.57 & 0.48 & 0.59 & 0.64 & 0.61 & 0.49 & 0.62 & 0.47 \\ 
GPT2-PPL & 0.60 & 0.48 &  0.51 & 0.28 & 0.52 & 0.46 & 0.21 & 0.34 & 0.47 & 0.19 & 0.44 \\ 

\midrule
\multicolumn{11}{l}{\textit{\textbf{(Question, answer) metrics}}} \\
QG & 0.63 & 0.58 & 0.51 & 0.60 & 0.61 & 0.56 & 0.59 & 0.50 & 0.56 & 0.64 & 0.48 \\ 
RankGen & 0.60 & 0.58 & 0.52  & \textbf{0.63} & 0.54 & 0.59 & \textbf{0.66} & 0.55 & 0.58 & \textbf{0.66} & 0.53 \\ 
BARTScore & 0.60 & 0.57 & 0.49  & 0.58 & 0.55 & 0.55 & 0.55 & 0.48 & 0.58 & 0.58 & 0.53 \\ 
\midrule
\multicolumn{11}{l}{\textit{\textbf{(Answer, evidence docs) metrics}}} \\
QAFactEval & - & 0.50 &  0.54 & - & - &  0.48 & - & - & \textbf{0.69} & - & - \\ 
\midrule
\multicolumn{11}{l}{\textit{\textbf{Learned metrics}}} \\
longformer & 0.67 & \textbf{0.62} & 0.59 & 0.60 & \textbf{0.62} & 0.56 & 0.62 & 0.65 & 0.63 & 0.63 & \textbf{0.63} \\ 
longformer (D) & - &  0.60 & \textbf{0.61} & - & - & 0.54 & - & - & 0.65 & - & -  \\ 
GPT3 curie & \textbf{0.69} &  0.55 & 0.59 & 0.60 & 0.51 & 0.45 & 0.53 & 0.55 & 0.58 & 0.56 &  0.51 \\ 
\midrule
Human & 0.80$^{\diamondsuit}$ & \multicolumn{2}{c}{0.73$^{\spadesuit}$} & - & - & - & - & - & - & - & - \\
\bottomrule
\end{tabular} 
\end{center}\vspace{-0.3em}
\caption{Accuracy of automatic metrics for imitating human judgments of overall answer preference, coherence, and factuality. \textbf{h/m} denotes comparisons between human-written answers and model-generated answers, while \textbf{m/m} denotes comparisons between pairs of model-generated answers. $^{\dagger}$These metrics are calculated on 109 pairs of comparisons, where comparisons of History are removed because there are only one answer available on the subreddit and hence no reference answer to compare. $^{\diamondsuit}$ We estimate the human performance with a pairwise agreement for two-way and three-way expert annotations. $^{\spadesuit}$ This pairwise agreement is reported by \textsc{WebGPT} \cite{nakano2021webgpt}, estimated on a subset of the data.}
\label{tab:overall_comparisons}
\vspace{-0.05in}
\end{table*}

\subsection{Evaluating automatic metrics}\label{sec:results}
\paragraph{Task}Each evaluation example consists of $\{(q, a_{1}, a_{2}, \hat{p})\}$, where $q$ is question, a pair of long-form answers $a_{1}$ and $a_{2}$, and $\hat{p}$ $\in$ \{$a_{1}$, $a_{2}$\} denotes the human preference of choosing answer $a_{1}$ or $a_{2}$. We report the accuracy of the metric preference $p_{i}$ against the gold human preference $\hat{p_{i}}$. We omit the evidence documents $d_{1}, d_{2}$ here for simplicity, but QAFactEval and longformer (D) metric take the evidence documents as additional input. 


\paragraph{Human preference data}
We compile human evaluations from previous studies \cite{Krishna2021HurdlesTP, nakano2021webgpt} and our expert annotations from Section \ref{sec:annotations}. See appendix~\ref{subsec:prev_human_eval} for descriptions of the models evaluated in these datasets as well as data statistics on the answers. Both prior studies present large-scale preference judgments of \textbf{overall} answer quality and smaller-scale judgments for two targeted aspects, \textbf{coherence} and \textbf{factuality}. In total, we look at 3,478 comparisons on overall answer quality, 854 comparisons on coherence, and 469 comparisons on factuality.  
As shown by our analysis of expert annotations (Section \ref{sec:annotations}), annotators can frequently disagree with each other.

\subsection{Results}
Table~\ref{tab:overall_comparisons} reports the accuracy of each metric at imitating human preference data. We report three baselines: \textbf{Random}, which randomly chooses one of the answers; \textbf{Always Human}, which prefers the human-written answer when available; and \textbf{Length}, which prefers the longer answer.\footnote{The \textbf{Length} baseline is inspired by prior findings in summarization \cite{Sun2019HowTC, Liu2022RevisitingTG} that \textbf{length} has a non-trivial impact in human preferences.}

\noindent\textbf{All metrics exhibit relatively low accuracies, falling substantially below estimated human agreement.} None of the metrics are robust across different types of input answer pairs. For instance, pretrained reference-based metrics such as BERTScore and BLEURT have low accuracy on \textsc{Hurdles} human vs.\ model data, which adds further evidence to the issues with ROUGE noted by~\citet{Krishna2021HurdlesTP}. Supervised metrics (Longformer and GPT-3) also struggle in this setting, despite outperforming all other metrics on overall rating in the other three data settings. While trained to imitate only overall rating, they achieve relatively strong accuracies on fine-grained ratings too, suggesting that they are correlated. 
\noindent\textbf{We observe spurious correlations with length for long-form answer evaluation.} Choosing the longer answer achieves higher accuracy than all unsupervised metrics for the \textsc{WebGPT} model vs.\ model comparison; the best performance on factuality for \textsc{Hurdles} human vs.\ model answer; and the second-highest accuracy on our expert data. On the other hand, when comparing \textsc{WebGPT} human vs.\ model answers, choosing a shorter answer would have been more beneficial for coherence evaluation (62\% of the time).
The ``strong'' performance of the length baseline displays the brittleness of all existing automatic metrics for LFQA.


\noindent\textbf{It is more feasible to model fine-grained answer aspects than overall answer quality.} The QAFactEval metric, designed for factuality, does indeed outperform all other metrics on factuality. However, the metric is limited in that it requires a set of input evidence documents, which may not always be available or reliable. For coherence, simpler metrics such as self-BLEU perform competitively, and we also find that our upper bound of always choosing the human answer performs strongly on coherence, suggesting that models struggle to generate coherent long-form answers. 

\noindent \textbf{Correlation of Automatic Metrics}
Given pairs of long-form answers of the comparison data, we measure how frequently two automatic metrics prefer the same answer (Figure \ref{fig:automatic_metrics_correlation}). We see a positive correlation among reference-based metrics (e.g., rouge and bertscore gives the same ranking for 63\% of the pairs), as well as the (question, answer) metrics (e.g. qg likelihood and bartscore). 

\input{figs/model_model_comparisons_metric_correlations.tex}

\section{Conclusion \& Future Work}

\noindent \textbf{Our study provides a unified evaluation benchmark for long-form answers, including new annotations from domain experts.} We present a new set of expert LFQA evaluations along with detailed justifications, and we also compile existing human annotations across different properties (overall preference, factuality, coherence) to facilitate future development of automatic LFQA metrics. 


\noindent  \textbf{Evaluation of long-form answers is a multi-faceted problem and thus should be more targeted.} Our expert justifications suggest that many aspects are considered when deciding which answer is better, some of which may be at odds with others (e.g. completeness vs. conciseness). This suggests that computing an ``overall'' score for answer quality is not meaningful, which is further supported by the limitations of metrics trained directly from overall preference judgments. Future work should look deeper into modelling frequent aspects mentioned by expert annotators, such as completeness and ease of understanding, perhaps by taking inspiration from evaluation methods that explicitly localize and categorize errors \cite{freitag-etal-2021-experts, goyal2022snac}.



\section*{Limitations}

We study a limited scope of long-form answers. The questions are either drawn from search queries or from community forums. In the real world, we will encounter many more diverse forms of long form question answering, such as answering questions in education or commercial settings. We only cover the English language, and thus our questions are topically limited to English-speaking culture. 

Our evaluation of long-form answers is stationary. Annotators are provided a pre-generated output from the model without being able to interact with the model over multiple rounds. A more interactive evaluation~\cite{Lee2022EvaluatingHM} of models is a great direction for future work.



\section*{Ethics Statement}

The expert annotation data collection protocol has been determined to be exempt from review by an IRB board. All data collected will be made publicly available under the MIT license. 

The data collection process did not require any information that can be used to uniquely identify individual workers. We examined the annotation data to make sure no such information or offensive content is present in questions or answers.

\section*{Acknowledgements}
MI and YS were partially supported by awards IIS-1955567 and IIS-2046248 from the National Science Foundation (NSF). FX is supported by a fellowship from UT Austin. We thank the WebGPT team, especially Jacob Hilton, for sharing their human evaluation data with us. We thank the expert annotators for participating in our human evaluation. We thank Jessy Li and members of the UT Austin NLP community for helpful discussion to improve the paper. Lastly, we thank the reviewers and meta reviewer of ACL community for helpful comments and feedback on the paper. 

\bibliography{anthology,custom}
\bibliographystyle{acl_natbib}
\newpage
\appendix

\section{Appendix}\label{sec:appendix}

\subsection{Related work on text generation evaluation}\label{subsec:related_work}
 Human and automatic evaluation for text generation is an active research area. We provide a brief overview here and direct the readers to recent surveys for more discussion \cite{Celikyilmaz2020EvaluationOT, gehrmann2022repairing}. Many tasks such as machine translation and summarization primarily rely on reference-based evaluation, with metrics such as BLEU \cite{papineni-etal-2002-bleu}, ROUGE \cite{lin-2004-rouge} and BERTScore \cite{zhang2019bertscore}. These metrics aim to measure similarities between generated text and reference text. For open-ended generation problems such as story generation, comparing the generated text with a single reference is not meaningful. Reference-based metrics which  instead measure the distributional similarity of model-generated and human-written texts have been proposed~\cite{Pillutla2021MAUVEMT}. There has also been work on reference-less metrics, which mostly measure a specific aspect of text. For instance, factuality metrics for summarization \cite{goyal-durrett-2020-evaluating, kryscinski-etal-2020-evaluating, barrantes2020adversarial, Laban2022SummaCRN} capture the relationship between source document and summary, without the need of a reference summary. Another line of work proposes automatic metrics which learn to emulate human judgements of generated text, using either gold human preference or synthetically generated data \cite{sellam2020bleurt, Zhong2022TowardsAU, Zhang2022FineDEvalFA}.


\subsection{Expert Annotation}\label{sec:expert_imp_details}
\begin{figure*}
    \centering
    \includegraphics[width=\linewidth]{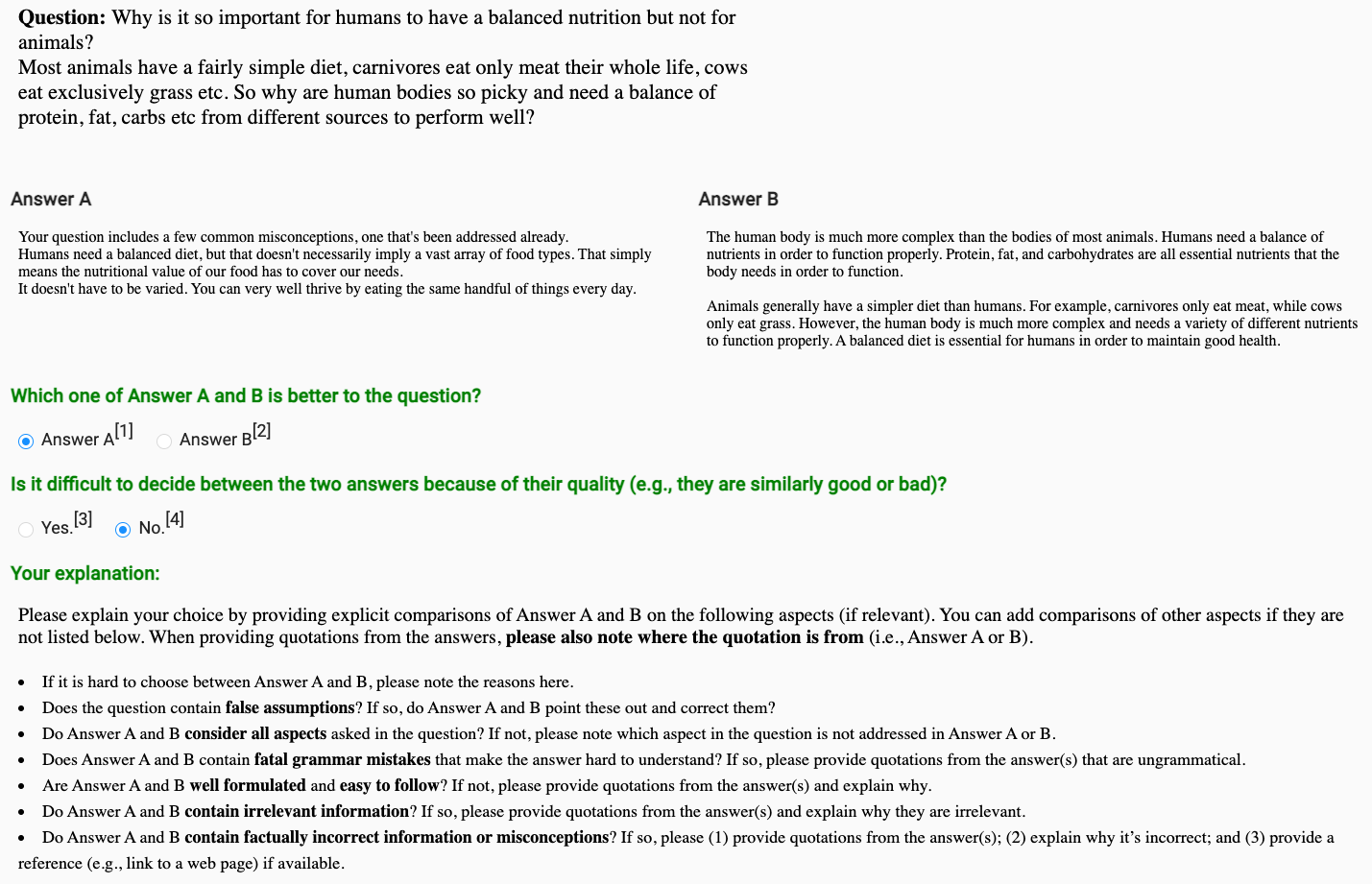}
    \caption{Screenshot of annotation interface for collecting expert evaluation.}
    \label{fig:annotation_interface}
\end{figure*}

\paragraph{Question clustering} Four domains (biology, physics, chemistry, and economics) are marked in the ELI5 posts (i.e., flairs), and two (tech/cs and law) are identified by using a dense passage retrieval~\cite{karpukhin-dpr} and KMeans from scikit-learn~\cite{scikit-learn}. Specifically, we use DPR to encode question of all posts whose flair is marked as \textit{others}. Then, we run KMeans to find two big groups of questions whose domains can be reliably marked as tech/cs and law.

\paragraph{Annotators} {Experts are hired based on their academic background and English proficiency. No other demographic and geographic restrictions were applied.} For each question domain, we aimed to hire three domain experts who have at least a bachelor's degree in the domain through a paid pilot study. Thirty-five potential experts participated in a paid pilot study with 5 question-answer {pairs}. We paid \$$3$ per question-answer set. {At the end, only 13 experts met the qualification requirements and were willing to continue because the task required substantive expertise as well as time and attention commitment.}

\begin{table*}[h]
\small
\centering
\begin{tabular}{@{}lrrrrrr@{}}
\toprule

& \multicolumn{2}{c}{Question} & \multicolumn{2}{c}{Model} & \multicolumn{2}{c}{Human} \\
Category  & Median    & Mean (std)       & Median  & Mean (std)      & Median  & Mean (std)      \\\midrule
Biology   & 20.50      & 49.40 (60.54)     & 74.00      & 75.70 (21.08)   & 56.00      & 79.20 (57.20)   \\
Physics   & 25.00        & 31.85 (18.70)    & 70.50    & 75.10 (27.06)   & 55.50    & 88.77 (82.91)   \\
Chemistry & 38.50      & 44.90 (29.13)     & 60.50    & 90.10 (92.79)   & 101.00     & 124.43 (77.59)  \\
Economics & 36.50      & 39.70 (30.93)     & 104.50   & 109.50 (50.75)  & 66.00      & 88.80 (93.21)   \\
Law       & 21.50      & 27.30 (19.38)     & 111.50   & 126.90 (75.31)  & 72.50    & 115.83 (146.48) \\
TechCS    & 21.50      & 35.10 (35.12)     & 91.00      & 94.90 (40.67)   & 105.00     & 112.43 (58.99)  \\
History   & 48.50      & 65.70 (57.87)     & 72.00      & 84.53 (58.24)   & 68.00      & 158.08 (168.97) \\\midrule
All       & 27.50      & 41.99 (41.01)    & 75.00      & 93.20 (59.93)   & 75.00      & 108.47 (106.56)\\\bottomrule
\end{tabular}
\caption{Statistics of the text length in our expert-annotated dataset. For each category, there are 20 questions. Each question has either a pair of human-written answers (H/H) or a pair of human-written and model-generated answers (H/M). The domain of history has 15 questions in the H/M setting and 5 in H/H. The other six domains have 10 questions in each setting. There are 140 questions, 205 human-written answers, and 75 model-generated answers.}
\label{tab:length_stats}
\end{table*}

\subsubsection{Justification Analysis}\label{subsec:full_masked_critique_anlaysis}

Data statistics of explanations collected are in Table \ref{tab:criqitue_stats}. Examples of explanation and extracted aspects in our manual analysis can be found in Table \ref{tab:explanation_example}.

\begin{table}
\small
\begin{center}
\begin{tabular}{@{}lrrr@{}}
\toprule
\textbf{Split} &\textbf{\# data} &\textbf{ Avg. \# word} & \textbf{Avg. \# span} \\
\midrule
Expert & 259 & 174 & 5\\
\textsc{WebGPT} & 292 & 46 & 3 \\
\bottomrule
\end{tabular} 
\end{center}\vspace{-0.3em}
\caption{Data statistics for computational analysis of free-form justifications. The span refers to the masked reference of candidate answer in the justifications.}
\label{tab:criqitue_stats}
\end{table}
\paragraph{Preprocessing} To construct the masked comments, we first preprocess the justifications such that all mentions of the answer entity is prepended with the word ``Answer'' (i.e. replacing ``Option A'', ``A'' with ``Answer A''). We then mask out any mentions of ``A'' and ``B'' in the comment. {We remove comments that do not contain answer entities after preprocessing, resulting in 259 (out of 260) expert comments and 292 (out of 305) \textsc{WebGPT} comments.}



\begin{table*}[ht!]
\small
\begin{tabular}{@{}p{2cm}p{2cm}p{11cm}@{}}
\toprule
\textbf{Aspect} & \textbf{Source} & \textbf{Comments}\\ \midrule
Factuality & Expert & [...] Answer B contains some \textbf{incorrect information} regarding the humans being more complex than animals and repeating same points twice. [...]\\ \midrule

Factuality & \textsc{WebGPT} & A claims pi bonds are the weakest, \textbf{which its sources don't state}, only calling them weaker than sigma bonds. A is also a little repetitive. B is much easier to follow and much simpler to understand. \\  \midrule

Easy to understand & Expert & [...] Of course, there is more to inflation than is provided by answer B, but it is concise, factual, and \textbf{easy to understand for someone that does not have a background in economics.}  [...]\\  \midrule

Relevance & Expert & For this question, Answer A is far better choice as it has accurate and scientific information \textbf{relevant to the question}. While answer B has irrelevant information by mentioning his personal experience of controlling the darkness which is totally over simplified statement. [...]\\ \midrule

Well-structured & Expert & [...] However, I decided that Answer B has provided more details and \textbf{is more well-structured compared to Answer A}.  [...] \\  \midrule

Completeness & Expert & For this question, answer B is better choice as it covers all aspects of the questions and explains the whole process with scientific facts.  While answer A contains \textbf{incomplete information} which cannot clear the doubts of reader. [...]\\ \midrule

Grammar & Expert & I believe option "A" is the better choice as it explains the meaning of a filibuster. \textbf{Option B lacks formal writing} and even states the words, "to shut him up". [...]\\  \midrule

Example & Expert & Both answers state the same information almost word for word. However, answer A provides \textbf{a clearer example for people who may not have experience in biology.} [...]\\  \midrule

Specificity & Expert & For this question, it is difficult to decide which is better option because both the answers are not up to the mark to clear the concept. Still, answer A seems better option as \textbf{it describes the process in detail} and mentioning  some harmones that involves in the process. [...]\\  \midrule

Conciseness & \textsc{WebGPT} & A is easier to follow, \textbf{much more concise}, and answers two possible interpretations of the question - the word's definition and the economic idea. B is \textbf{overly detailed} and needlessly argues with the use of austerity. A is much better. 

\\\bottomrule
\end{tabular} \vspace{-0.3em}
\caption{Free-form justifications written by experts and their corresponding aspects.}
\label{tab:explanation_example}
\end{table*}

\begin{table*}
\tiny
\setlength{\tabcolsep}{3pt}
\begin{center}
\begin{tabular}{lp{1.5cm}p{3.3cm}p{3.3cm}p{3cm}p{3cm}}
\toprule
\textbf{Domain} & \textbf{Question} & \textbf{Answer A}& \textbf{Answer B} & \textbf{Expert 1} & \textbf{Expert 2} \\
\midrule
Physics & How does a speaker vibrate at multiple frequencies simultaneously to deliver sounds to our ears? & This has been asked many times and the answer is they don’t. If you listen to the song being played live on purely acoustic instruments even though they are being played separately and emitting their own frequencies, what you hear (and by extension, what a microphone captures) at any given time is just ONE frequency that’s the “sum” of all the others combined. A speaker is just a reverse microphone. & Imagine an ocean with a consistent wave. It flows up and down, with equal distance between the two waves at any time.Now imagine I push a larger, shorter wave into this ocean. The two waves will collide, resulting in some new wave pattern. This new wave pattern is a combination of those two waves.Speakers work similarly. If I combine two soundwaves, I get a new combination wave that sounds different. & \textbf{Justifications}: In technical terms, ocean waves stated in answer \textcolor{cyan}{B} are transverse waves and sound waves are longitudinal waves. [...] So in terms of \textbf{technical approach and considering overall accuracy} answer \textcolor{purple}{A} is selected. \newline \textbf{Preference}: \textcolor{purple}{A} \newline \textbf{Aspects}: Factuality &   \textbf{Justifications}: I go with answer \textcolor{cyan}{B} because \textbf{I like the analogy with the ocean waves}, and due to how visual the explanation is it is easier to understand in my opinion. [...] \newline \textbf{Preference}: \textcolor{cyan}{B} \newline \textbf{Aspects}: Example, Easy-to-follow \\
\midrule
Economics & What are toll roads or "turnpikes" used for? & There is a toll road near my house. It was built by a private company on private land. It represents a significant short-cut between two state-built roads. The company built the road, and collects tolls for its use, in an effort to make money.So, in this case, the toll goes to repay the initial investment to acquire the land and build the road. It also goes toward providing snow removal, maintenance, and repair. & The money from the tolls usually goes towards maintaining the road, and/or is like a travel tax that the state can use for other projects. Sometimes tolls are only there to cover the cost of constructing the road and then once it's paid off they take the toll off (that's less common though). & \textbf{Justifications}: Answer \textcolor{purple}{A} better answers the original question, although both could have gone into more detail. [...] Answer \textcolor{purple}{A} does \textbf{address the latter point}. \newline \textbf{Preference}: \textcolor{purple}{A} \newline \textbf{Aspects}: Completeness &  \textbf{Justifications}: [...]Answer \textcolor{purple}{A} explains only a single use case of the use of tolls \textbf{while the question demands a broader answer which is mentioned in Answer \textcolor{cyan}{B}.} [...] \textcolor{purple}{A} contains \textbf{irrelevant information} [...] \newline \textbf{Preference}: \textcolor{cyan}{B} \newline \textbf{Aspects}: Completeness, Relevance \\
\midrule 
Biology & How are sperms made in the body? More specifically, how does a Sperm become X or Y sperm. & Not going too much into the scientificy details, mainly because I don't remember them from high school  2 decades ago, but basically the cell division that occurs to form the reproductive cells (sperm/eggs) in humans is not the same as the cell division for none reproductive cells. When the "normal" cells split, they create complete copies of each chromosome pair (your DNA is made of pairs of each chromosome. One comes from the father, one from the mother), so the child cells end up with a complete set of DNA. Reproductive cells split the chromosome pairs. The child cells only receive one chromosome from each pair. In the case of the sex chromosome pair, a male has an XY pair and a female has an XX pair. So when a male creates reproductive cells (sperm), one sperm will receive the X chromosome and the other will have the Y chromosome. & When most cells divide, they copy themselves once then split once, so both copied cells have X *and* Y. Sperm basically come from 1 copy but 2 splits, so you get one X sperm and one Y sperm from the second split. & \textbf{Comments}: Both answers state the same information almost word for word. However, answer \textcolor{purple}{A} provides \textbf{a clearer example for people who may not have experience in biology.} [...] \newline \textbf{Preference}: \textcolor{purple}{A}   \newline \textbf{Aspects}: Example & \textbf{Comments}: Answer \textcolor{cyan}{B} \textbf{doesn't distinguish} between men and women which is pertinent in this question. Answer \textcolor{cyan}{B} \textbf{lacks detail to make the answer clear.} [...] Answer \textcolor{purple}{A} has \textbf{a better flow, is more comprehensive} and better answers the question." \newline \textbf{Preference}: \textcolor{purple}{A} \newline \textbf{Aspects}: Detailed, Easy to follow \\
\bottomrule
\end{tabular} 
\end{center}\vspace{-0.3em}
\caption{Example annotations by domain experts comparing long-form answers, either generated from GPT3 or human written, showing their preferences, free-form justifications and aspects. The first two examples illustrate that experts disagree with each other because they value different aspects.}
\label{tab:comment_examples}
\end{table*}

\begin{table*}
\footnotesize
\begin{center}
\begin{tabular}{lrrr}
\toprule
\textbf{Data} &\textbf{\# data}& \textbf{\# non-tie data} & \textbf{Aspect} \\
\midrule
\textsc{Hurdles} (human v.s. model) & 486 / 214 / 194  & 419 / 164 / 151 & Overall / Coherence / Factuality  \\   
\textsc{Hurdles} (model v.s. model) & 521 / 262 / 260  & 370 / 195 / 169 & Overall / Coherence / Factuality  \\   
\textsc{WebGPT} (human v.s. model) & 761 / 761 / 590  & 637 / 496 / 149 & Overall / Coherence / Factuality  \\ 
\textsc{WebGPT} (model v.s. model) & 17,598  & 13,065 & Overall \\
\bottomrule
\end{tabular} 
\end{center}\vspace{-0.3em}
\caption{Data Statistics for human \textbf{comparison} evaluation data for each aspect. In all studies, overall score was mandatory but coherence / factuality scores were optional and hence the number of evaluation data available varies among different aspects. All human evaluation data is one-way annotated.}
\label{tab:comparison_dataset_stats}
\end{table*}

\subsection{Previously Collected Human Evaluation Data}\label{subsec:prev_human_eval}

Dataset statistics is shown in Table \ref{tab:comparison_dataset_stats}. We group the comparisons by whether they are (model-generated answers v.s. human-written answers) or (model-generated answers v.s. model-generated answers), and present overall statistics. The model-generated answers include four different set-ups from \textsc{Hurdles} (combination of nucleus sampling p=\{0.6, 0.9\}, and generation conditioning on \{predicted, random\} passages) and three different set-ups from \textsc{WebGPT}. The human-written answers are gold answers from the ELI5 subreddit for comparison with \textsc{Hurdles} answers, and human demonstrations for \textsc{WebGPT} answers.

\subsubsection{LFQA systems}
We describe the different LFQA systems developed by prior works, which are included in comparisons used for evaluating automatic metrics in Section \ref{sec:metrics}.

\paragraph{\textsc{Hurdles}} \citet{Krishna2021HurdlesTP} presented a state-of-the-art LFQA system which includes a passage retriever \cite{realm} and an answer generation model \cite{Roy2021EfficientCS}.

\paragraph{\textsc{WebGPT}} \citet{nakano2021webgpt} proposed to fine-tune GPT-3 \cite{brown2020language} to interact with a search engine and compose long-form answers based on the information found. The generated answers also contain a set of reference documents found online.

\subsubsection{Evaluation aspects}
We describe the different evaluation aspects conducted by prior human evaluation.

\paragraph{Overall} \citet{Krishna2021HurdlesTP} phrased the question as ``Which generation answered the question better / was more relevant to the question?'' while \citet{nakano2021webgpt} developed detailed instructions with intermediate steps for comparing two answers, and dedicated an overall rating, phrased as ``how useful the answer would be to the person asking the question, all things considered''. 

\paragraph{Coherence} \citet{Krishna2021HurdlesTP} asked the human evaluators to choose the more coherent answer and listed repetition as a trait of incoherence.\footnote{The wording was (which answer) ``was more coherent / had less repetition''.} In \citet{nakano2021webgpt}, the instruction for coherence evaluation focuses on whether the answer makes sense, is easy to follow and is in a logical order.

\paragraph{Factuality} \citet{Krishna2021HurdlesTP} instructed human evaluators to judge factual correctness of answers, with no accompanying evidence documents but permission to use search engine over Wikipedia articles. In \citet{nakano2021webgpt}, the evaluation of factuality is focused on whether the generated answer could be entailed by the evidence documents and that it doesn't hallucinate unsupported fact. Note that ``faithfulness'' to the evidence articles is a different notion from the ``correctness'' of the answer, as the evidence articles might not always be correct or up-to-date \cite{Gao2022AttributedTG}.

\subsubsection{Example of comments mentioning different aspects for Section \ref{subsec:manual-analysis}}\label{appendix:example-for-sec3.3}

See Table \ref{tab:takeaway}.

\begin{table*}[]
\scriptsize
\resizebox{0.98\textwidth}{!}{%
\begin{tabular}{@{}c@{\hspace{1.3ex}}p{2cm}@{\hspace{2ex}}p{3.7cm}@{\hspace{2ex}}p{3.7cm}@{\hspace{2ex}}p{3.2cm}@{}}
\toprule
Aspect &
  Question &
  Answer A &
  Answer B &
  Comment \\\midrule
\multirow{1}{*}{\rotatebox[origin=c]{90}{Factuality }} &
  Does your body absorb all blood clots? How does it do it? Does location affect it? &
  (human) Kind of. Your blood will dissolve clots over time. I've got a clotting disorder and need to take thinners to make that happen faster, but for most they break down clots naturally settlement on how much blood flow there is to an area and if there's any muscle action to help break them up. [...] &
  (machine) No, your body does not absorb all blood clots. Clots are usually broken down by the body's natural clotting mechanism, but some clots may persist and cause problems. Location can affect how quickly a clot is broken down. For example, clots in the veins are usually broken down more slowly than clots in the arteries. &
  (expert) {[}...{]} Answer A mentioning some inaccurate and irrelevant information, like role of blood thinner which is not correct as they cannot break down the clots. {[}...{]}\\\midrule
  \multirow{1}{*}{\rotatebox[origin=c]{90}{Completeness}} & Why do people go bald on the top of their head and still have some on the sides and not vice-versa? & (machine) There are a few reasons why people go bald on the top of their head and still have some hair on the sides. One reason is that the hair follicles on the top of the head are more sensitive to a hormone called dihydrotestosterone (DHT). DHT is [...] found in both men and women. [...] Another reason is that the hair on the sides and back of the head is not as exposed to the sun and other environmental factors, [...] & (human) Seems unsettled but here’s a theory: “The most recent hypothesis suggests that the hair loss process begins during puberty, when growth of the skull and the muscles in the forehead and neck increases the tension in a tight band of tissue stretching over the top of the head. The more DHT (a type of testosterone) there is, the more these muscles contract, which causes inflammation and brings in more DHT. [...] & (expert) It is difficult to decide but I believe answer A is better choice as it describes both the harmonal and environmental causes behind the baldness of top. And Answer B only focuses on one theory which might not be 100 percent accurate. [...] According to research, baldness is due to genes. In 95 percent cases, balding is due to androgenetic alopecia, more commonly known as male pattern baldness, which is hereditary condition. [...] \\
  \midrule
  \multirow{1}{*}{\rotatebox[origin=c]{90}{Specificity}} & "Why do solar systems look mostly 'flat'? I mean like, \href{https://i.imgur.com/uG6tHHa.gif}{it seems as though all the planets are rotating on roughly the same axis.} Why is there little to no variation? " & (machine) The solar system is not actually a perfect plane, but it is well on its way to an ideal stage. Pluto actually lies at an awkward incline, due to its low gravity and isolated orbit. The asteroid belt, in its most oblique locations, also inclines at about 20 degrees from the main plane. Aside from these few exceptions though, the power of angular momentum has left the planets of our solar system in a tidy, predictable plane.

  The reason for this is that [...] & (human) The solar system is flat due to the law of conservation of total angular momentum. This law states that the total amount of spinning in any isolated system will always remain constant. This is what caused the solar system to flatten as time went on and becoming a flat disc. & (WebGPT) B is overly brief and simplistic, not explaining its concepts well enough for them to be understandable. A is longer, but it is very useful, interesting, and detailed. It answers the question about the solar system being flat thoroughly and coherently. It is a much better answer than B. \\\midrule
  \multirow{1}{*}{\rotatebox[origin=c]{90}{Use of examples}} & (human) If under anesthesia do you get your sleep needed for the day? & No. The unconsciousness induced by both IV anesthetics and inhaled anesthetics is very different from restorative sleep. [...] True ELI5: Sleep = charging your phone on airplane mode. Anesthesia = turning your phone off but not charging it. [...] & (machine) There is no clear consensus on whether or not people under anesthesia get the sleep they need for the day. Some studies suggest that people under anesthesia do get some sleep, but it is not clear how deep or restful this sleep is. Other studies suggest that people under anesthesia do not get any meaningful sleep at all. Therefore, it is difficult to say definitively whether or not people under anesthesia get the sleep they need for the day. & (expert) For this question, answer A is better option because it has proper explanation of relevant information which is being asked in the question. [...] In answer A, relevant detail of the answer is explained using the example of equating sleep to charging the phone and anesthesia to turning it off but not charging which clear the idea about this concept. [...] \\
  \bottomrule
\end{tabular}%
}
\caption{Examples of some of the aspects that occur in the manual analysis described in Section \ref{subsec:manual-analysis}. }
\label{tab:takeaway}
\end{table*}

\subsection{Automatic Metric Implementation Details}\label{subsec:implementation_details}

{Length statistics of the answers evaluated in \ref{sec:automatic_metrics} are reported in Table \ref{tab:human_data_length_stats}. We truncate the input if it exceeds the context window for the model. Less than 5\% of the comparison data are truncated.}

\paragraph{ROUGE-L} For each answer, we calculate ROUGE-L against the set of reference answers from ELI5 and use the maximal ROUGE-L. 

\paragraph{BERTScore} We use the default \texttt{roberta-large} model for English\footnote{\url{https://github.com/Tiiiger/bert_score}} and report the maximal F1 BERT score against the set of reference answers.

\paragraph{BLEURT} We use the \texttt{BLEURT-20} checkpoint as recommended and report the maximal BLEURT score against the set of reference answers.

\paragraph{Self-BLEU} We calculate Self-BLEU by regarding one sentence as hypothesis and all others in the same answer paragraph as reference. We report self-BLEU-5 as a measure of coherence.

\paragraph{Length} We use the Stanza toolkit \cite{qi-etal-2020-stanza} for word tokenization.

\paragraph{QG Likelihood} Given a question $q$ and an answer paragraph $a$, we estimate $p(q|a)$ by computing the average log-likelihood of the question tokens conditioned on the passage using T0. Following previous work \cite{sachan2022improving}, we append a natural language instruction \textit{``Which question does this passage answer?''} to the answer, denoted as $a^\prime$.
$$\text{log}p(q|a) = \frac{1}{|\boldsymbol q|}\sum_{t}\text{log}p(q_{t}|\boldsymbol q_{<t}, a^\prime; \Theta)$$
where $\Theta$ denotes the parameter of the language model and $|\boldsymbol q|$ denotes the number of tokens in the question.

\paragraph{BARTScore} We use the BART model finetuned on the CNN/DM dataset (\texttt{facebook/bart-large-cnn}).

\paragraph{RankGen} Given a question $q$ and an answer paragraph $a$, we first encode them through the RankGen encoder, which projects them to fixed-size vectors $(\boldsymbol q, \boldsymbol a)$. We then determine their relevance by calculating the dot product between the two vectors $\boldsymbol q \cdot \boldsymbol a$. We use the T5-XXL (11B) encoder trained on both in-book negative and generative negatives.

\paragraph{QAFactEval} QAFactEval \cite{fabbri-etal-2022-qafacteval} is a recently proposed QA-based metric that has shown superior performane on several summarization factuality benchmark \cite{Laban2022SummaCRN, maynez-etal-2020-faithfulness}. The pipeline is carefully chosen from extensive experiments on various combinations of components in the QA-based metrics. The final pipeline consists of (1) NP from $S$ as $Ans(S)$ (2) BART-large \cite{lewis-etal-2020-bart} as $Q_{G}$ (3) Electra-large \cite{clark2020electra} as $Q_{A}$ and (4) learned metrics \textbf{LERC} \cite{chen-etal-2020-mocha} as $Sim(p_{i}, s_{i})$. They further include an answerability classification module to determine if the question is answerable given the document $D$. We report the \textbf{LERC}, which uses the learned metrics to compare $Ans_{S}$ and $Ans_{D}$(a) and shows better performance compared to other metrics in our initial experiments.

\begin{table}[]
\centering
\resizebox{\columnwidth}{!}{%
\begin{tabular}{@{}lrrr@{}}
\toprule
Category & \# QA pairs & Consistency & Accuracy \\ \midrule
Biology & 11 & 100\% & 82\% \\
Physics & 13 & 100\% & 62\% \\
Economics & 12 & 92\% & 83\% \\
History & 13 & 100\% & 100\% \\ \bottomrule
\end{tabular}%
}
\caption{Performance of 2 shot question answer evaluation using GPT3 text-davinci-003. Consistency reports the percentage of the model generate the same preferred answer across three API calls. Accuracy compares the majority votes among the three API calls against the human preference.}
\label{tab:2shot_003}
\end{table}

\begin{table}
\small
\begin{center}
\begin{tabular}{lrr}
\toprule
\textbf{Split} &\textbf{\# data} &\textbf{\# non-tie data}\\
 \midrule
 train  & 12,318 & 9,153 \\
 dev & 2,640 & 1,989  \\ 
 test & 2,640 & 1,923 \\ 
 \midrule
 total & 17,598 & 13,065 \\
\bottomrule
\end{tabular} 
\end{center}\vspace{-0.3em}
\caption{Data statistics for human preference data used to train and evaluate the learned metric (Section \ref{sec:reward_model}). We collapse human rating such that the answer preferred is assigned score 1 and the other 0. \citet{nakano2021webgpt} included tie data and assign them 50\% soft labels, and excluded them from evaluation. However, we didn't find them beneficial for model training and hence removed them from both training and valuation.}
\label{tab:rm_data_stats}
\end{table}

\begin{table}
\small
\begin{center}
\begin{tabular}{@{}lccccc@{}}
\toprule
\textbf{Answer Type} & \# answer & $|q|$ & $|a|$ & $|d|$ & $|j|$\\
\midrule
\textsc{WebGPT human}  & 254 & 35 & 112 & 264  & \multirow{2}{*}{46} \\  
\textsc{WebGPT model}  & 6,095 & 35 & 137 & 328 \\ 
\textsc{Hurdles human} & 442 & 17 & 300 & - & \multirow{2}{*}{-} \\ 
\textsc{Hurdles model} & 1,135& 17 & 182 & - \\
\textsc{Expert human} & 205 & 42 & 108 & - & \multirow{2}{*}{176}\\ 
\textsc{Expert model} & 75 & 42 & 93 & - \\
\bottomrule
\end{tabular} 
\end{center}\vspace{-0.3em}
\caption{Data statistics of answers compared in the human evaluation data. The number of comparison data can be found in Table~\ref{tab:overall_comparisons}. $|q|$, $|a|$ ,$|d|$ and $|j|$ represent the average number of words for question, answer paragraph, retrieved documents and justification. For WebGPT, justifications are only on a subset of comparison data. WebGPT and expert annotation data take both the title and the description of the reddit post as question following \cite{nakano2021webgpt}, whereas Hurdles data only considers the title as question (hence shorter $|q|$).}
\label{tab:human_data_length_stats}
\end{table}

\subsubsection{Learned Metrics}
\label{subsec:finetuningdetails}
We use \verb|pytorch-transformers| \citet{Wolf2019HuggingFacesTS} to implement our models. We use Quadro RTX 8000 GPUs to train our model. 

\paragraph{Longformer} We use \texttt{longformer-base}, consisting of ~149M parameters. The training batch size is set to 16, with the initial learning rate as $1e-5$. We used AdamW optimizer and a linear learning rate schedule. We train the model for 5 epochs and report the result of the checkpoint with best validation accuracy. The training takes less than 5 hours with 4 GPUs.

\paragraph{GPT3}
We use the API to fine-tune the model with a batch size of 64 and a learning rate multiplier 0.05 for six epochs. {Fine-tuning text-curie001 model for each epoch on OpenAI cost \$11. We did not use the larger \texttt{text-davinci-002} model, which would have cost \$110 per epoch.}


\subsubsection{GPT-3 Two-shot}
We conduct a pilot study on prompting GPT3 \texttt{text-davinci-003} for the pair-wise answer evaluation task on a subset of our expert annotation data.

For each domain that has multiple experts (i.e., biology, physics, economics, and history), we evaluate on the questions for which all experts agreed on the label of the preferred answer. We randomly choose two question-answer sets as the in-context example  and prompt the model on the rest of the question-answer sets. The prompt has the following format:

\begin{quote}
\small
    QUESTION: $q$

    ANSWER1: $a_{1}$

    ANSWER2: $a_{2}$

    TASK: Choose the better answer.

    BETTER ANSWER: ANSWER1 (or ANSWER2) is better.
\end{quote}

\noindent For each question-answer set, we sample three times  with top $p$ = 1 and temperature = 0.7 to evaluate model's consistency. The results are reported in Table \ref{tab:2shot_003}.

Results are report in Table \ref{tab:2shot_003}. The model is mostly self-consistent.Model also aligns with human on this small set of data where human have perfect agreement with each other, model aligns with human performance, despite variance across different domains. We leave further investigation on utilizing large language model for automatic evaluation on long-form question answering to future work.

\end{document}